%% file: nips2011.tex
\documentclass{article} % For LaTeX2e
\usepackage{nips11submit_e,times}
\input defs.tex

\def\myxi{x}
\def\onehalf{\frac{1}{2}}
\newtheorem{theorem}{Theorem}[section]

\usepackage{graphicx}
\usepackage{epstopdf}
\usepackage{color}
\usepackage{psfrag,pdfsync}
\usepackage{subfigure}

\title{Large-Scale Sparse Principal Component Analysis \\ with Application to Text Data}

\author{
Youwei Zhang\\
Department of Electrical Engineering and Computer Sciences\\
University of California, Berkeley\\
Berkeley, CA 94720 \\
\texttt{zyw@eecs.berkeley.edu} \\
\And
Laurent El Ghaoui\\
Department of Electrical Engineering and Computer Sciences\\
University of California, Berkeley\\
Berkeley, CA 94720 \\
\texttt{elghaoui@eecs.berkeley.edu} \\
}

\nipsfinalcopy % Uncomment for camera-ready version

\begin{document}

\maketitle

\begin{abstract}
Sparse PCA provides a linear combination of small number of features that maximizes variance across data. Although Sparse PCA has apparent advantages compared to PCA, such as better interpretability, it is generally thought to be computationally much more expensive. In this paper, we demonstrate the surprising fact that sparse PCA can be easier than PCA in practice, and that it can be reliably applied to very large data sets. This comes from a rigorous feature elimination  pre-processing result, coupled with the favorable fact that features in real-life data typically have exponentially decreasing variances, which allows for many features to be eliminated. We introduce a fast block coordinate ascent algorithm with much better computational complexity than the existing first-order ones. We provide  experimental results obtained on text corpora involving millions of documents and hundreds of thousands of features. These results illustrate how Sparse PCA can help organize a large corpus of text data in a user-interpretable way, providing an attractive alternative approach to topic models.  
\end{abstract}

\section{Introduction}
The sparse Principal Component Analysis (Sparse PCA) problem is a variant of the classical PCA problem, which accomplishes a trade-off between the explained variance along a normalized vector, and the number of non-zero components of that vector. 

Sparse PCA not only brings better interpretation~\cite{AEJL:07}, but also provides statistical regularization \cite{Amin08} when the number of samples is less than the number of features. Various researchers have proposed different formulations and algorithms for this problem, ranging from ad-hoc methods such as factor rotation techniques \cite{Joll95} and simple thresholding  \cite{cadi95}, to greedy algorithms \cite{Mogh06b,  dAsp08b}. Other algorithms include SCoTLASS by \cite{Joll03},  SPCA by \cite{Zou04}, the regularized SVD method by \cite{SH:08} and the generalized power method by \cite{Jour08}. These algorithms are based on non-convex formulations, and may only converge to a local optimum. The $\ell_1$-norm based semidefinite relaxation DSPCA, as introduced in~\cite{AEJL:07}, does guarantee global convergence and as such, is an attractive alternative to local methods. In fact, it has been shown in \cite{AEJL:07, Amin08, ZAE:11} that simple ad-hoc methods, and the greedy, SCoTLASS and SPCA algorithms, often underperform DSPCA. However, the first-order algorithm for solving DSPCA, as developed in~\cite{AEJL:07}, has a computational complexity of $O(n^4\sqrt{\log n})$, with $n$ the number of features, which is too high for many large-scale data sets. At first glance, this complexity estimate indicates that solving sparse PCA is much more expensive than PCA, since we can compute one principal component with a complexity of $O(n^2)$. 

In this paper we show that solving DSPCA is in fact computationally easier than PCA, and hence can be applied to very large-scale data sets. To achieve that, we first view DSPCA as an approximation to a harder, cardinality-constrained optimization problem.  Based on that formulation, we describe a safe feature elimination method for that problem, which leads to an often important reduction in problem size, prior to solving the problem. Then we develop a block coordinate ascent algorithm, with a computational complexity of $O(n^3)$ to solve DSPCA, which is much faster than the first-order algorithm proposed in~\cite{AEJL:07}. Finally, we observe that real data sets typically allow for a dramatic reduction in problem size as afforded by our safe feature elimination result. Now the comparison between sparse PCA and PCA becomes $O(\hat n^3)$ v.s. $O(n^2)$ with $\hat n \ll n$, which can make sparse PCA surprisingly easier than PCA.

In Section 2, we review the $\ell_1$-norm based DSPCA formulation, and relate it to an approximation to the $\ell_0$-norm based formulation and highlight the safe feature elimination mechanism as a powerful pre-processing technique. We use Section 3 to present our fast block coordinate ascent algorithm. Finally, in Section 4, we demonstrate the efficiency of our approach on two large data sets, each one containing more than 100,000 features.

\paragraph{Notation.} ${\cal R}(Y)$ denotes the range of matrix $Y$, and $Y^\dagger$ its pseudo-inverse.  The notation $\log$ refers to the extended-value function, with $\log \myxi = -\infty$ if $\myxi \le 0$.

\section{Safe Feature Elimination}
\paragraph{Primal problem.}
Given a $n \times n$ positive-semidefinite matrix $\Sigma$, the ``sparse PCA'' problem introduced in~\cite{AEJL:07} is :
\begin{equation}\label{eq:primal-pb}
\phi = \max_Z \: \Tr \Sigma Z - \lambda \|Z\|_1 ~:~ Z \succeq 0, \;\; \Tr Z = 1
\end{equation}
where $\lambda \ge 0$ is a parameter encouraging sparsity. Without loss of generality we may assume that $\Sigma \succ 0$. 

Problem~(\ref{eq:primal-pb}) is in fact a relaxation to a PCA problem with a penalty on the cardinality of the variable:
\begin{equation} \label{eq:primal-pb-card}
\psi = \max_x \: x^T \Sigma x - \lambda \|x\|_0  ~:~  \|x\|_2 = 1 
\end{equation}
Where $\|x\|_0$ denotes the cardinality (number of non-zero elemements) in $x$.  This can be seen by first writing problem~(\ref{eq:primal-pb-card}) as: 
\[
\max_Z \: \Tr \Sigma Z - \lambda \sqrt{ \|Z\|_0} ~:~ Z \succeq 0, \;\; \Tr Z = 1 , \mathop{\bf Rank}(Z) =1
\]
where $\|Z\|_0$ is the cardinality (number of non-zero elements) of $Z$.  Since $\|Z\|_1 \le \sqrt{\|Z\|_0} \: \|Z\|_F = \sqrt{\|Z\|_0}$, we obtain the relaxation
\[
\max_Z \: \Tr \Sigma Z - \lambda \|Z\|_1 ~:~ Z \succeq 0, \Tr Z =1, \mathop{\bf Rank}(Z) =1
\]
Further drop the rank constraint, leading to problem~(\ref{eq:primal-pb}).

By viewing problem~(\ref{eq:primal-pb}) as a convex approximation to the non-convex problem~(\ref{eq:primal-pb-card}), we can leverage the safe feature elimination theorem first presented in~\cite{dAsp08b, Elg:06} for problem~(\ref{eq:primal-pb-card}):
\begin{theorem}%[Thresholded Expression for Problem~\ref{eq:primal-pb-card}]
\label{SAFE}
Let $\Sigma = A^T A$, where $A = (a_1, \ldots, a_n) \in \reals^{m \times n}$. We have
\[
\psi  = \max_{\|\xi\|_2 = 1}  \sum^n_{i=1} ((a_i^T \xi)^2 -\lambda)_+ .
\]
An optimal non-zero pattern corresponds to indices $i$ with $\lambda < (a_i^T \xi)^2$ at optimum.
\end{theorem}
We observe that the $i$-th feature is absent at optimum if  $(a_i^T \xi)^2 \le \lambda \mbox{ for every } \xi, \;\; \|\xi\|_2 = 1$. Hence, we can {\em safely} remove feature $i \in \{1,\ldots,n\}$ if 
\begin{equation}
\label{safe}
\Sigma_{ii} = a_i^T a_i < \lambda 
\end{equation}
A few remarks are in order. First, if we are interested in solving problem~(\ref{eq:primal-pb}) as a relaxation to problem~(\ref{eq:primal-pb-card}), we first calculate and rank all the feature variances, which takes $O(nm)$ and $O(n\log (n))$ respectively. Then we can safely eliminate any feature with variance less than $\lambda$. Second, the elimination criterion above is conservative. However, when looking for extremely sparse solutions, applying this safe feature elimination test with a large $\lambda$ can dramatically reduce problem size and lead to huge computational savings, as will be demonstrated empirically in Section 4. Third, in practice, when PCA is performed on large data sets, some similar variance-based criteria is routinely employed to bring problem sizes down to a manageable level. This purely heuristic practice has a rigorous interpretation in the context of \emph{sparse} PCA, as the above theorem states explicitly the features that can be safely discarded.

\section{Block Coordinate Ascent Algorithm}

The first-order algorithm developed in~\cite{AEJL:07} to solve problem~(\ref{eq:primal-pb}) has a computational complexity of  $O(n^4\sqrt{\log n})$. With a theoretical convergence rate of $O(\frac{1}{\epsilon})$, the DSPCA algorithm does not converge fast in practice. In this section, we develop a block coordinate ascent algorithm with better dependence on problem size ($O(n^3)$), that in practice converges much faster. 

\paragraph{Failure of a direct method.} We seek to apply a ``row-by-row'' algorithm by which we update each row/column pair, one at a time. This algorithm appeared in the specific context of sparse covariance estimation in \cite{BEA:08}, and extended to a large class of SDPs in \cite{WGMS:09}. Precisely, it applies to problems of the form
\begin{equation}\label{eq:general-form}
\min_X \: f(X) - \beta \log \det X ~:~ L \le X \le U, \;\; X \succ 0,
\end{equation}
where $X = X^T$ is a $n \times n$ matrix variable, $L,U$ impose component-wise bounds on $X$, $f$ is convex, and $\beta >0$.

However, if we try to update the row/columns of $Z$ in problem~(\ref{eq:primal-pb}), the trace constraint will imply that we never modify the diagonal elements of $Z$. Indeed at each step, we update only one diagonal element, and it is entirely fixed given all the other diagonal elements.  The row-by-row algorithm does not directly work in that case, nor in general for SDPs with equality constraints.  The authors in \cite{WGMS:09} propose an augmented Lagrangian method to deal with such constraints, with a complication due to the choice of appropriate penalty parameters.  In our case, we can apply a technique resembling the augmented Lagrangian technique, without this added complication.  This is due to the homogeneous nature of the objective function and of the conic constraint. Thanks to the feature elimination result (Thm.~\ref{SAFE}), we can always assume without loss of generality that $\lambda < \sigma_{\rm min}^2 :=\min_{1 \le i \le n} \Sigma_{ii}$.

\paragraph{Direct augmented Lagrangian technique.}
We can express problem~(\ref{eq:primal-pb}) as
\begin{equation}\label{eq:primal-pb-aug}
\frac{1}{2}\phi^2 = \max_X \: \Tr \Sigma X - \lambda \|X\|_1 - \frac{1}{2} (\Tr X)^2 ~:~ X \succeq 0 .
\end{equation}
This expression results from the change of variable $X = \gamma Z$, with $\Tr Z = 1$, and $\gamma \ge 0$.  Optimizing over $\gamma \ge 0$, and exploiting $\phi > 0$ (which comes from our assumption that $\lambda < \sigma_{\rm min}^2$), leads to the result, with the optimal scaling factor $\gamma$ equal to $\phi$. An optimal solution $Z^\ast$ to~(\ref{eq:primal-pb}) can be obtained from an optimal solution $X^\ast$ to the above, via $Z^\ast = X^\ast/\phi$. (In fact, we have $Z^\ast = X^\ast/\Tr(X^\ast)$.)

To apply the row-by-row method to the above problem, we need to consider a variant of it, with a strictly convex objective.  That is, we address the problem
\begin{equation}\label{eq:primal-pb-aug-pen}
\max_X \: \Tr \Sigma X - \lambda \|X\|_1 - \frac{1}{2} (\Tr X)^2 + \beta \log \det X, ~:~ X \succ 0,
\end{equation}
where $\beta>0$ is a penalty parameter.  SDP theory ensures that if $\beta = \epsilon/n$, then a solution to the above problem is $\epsilon$-suboptimal for the original problem \cite{BV:04}. 

\paragraph{Optimizing over one row/column.} Without loss of generality, we consider the problem of updating the last row/column of the matrix variable $X$.  Partition the latter and the covariance matrix $S$ as
\[
X = \left( \begin{array}{cc} Y & y \\ y^T & \myxi \end{array} \right), \;\;
\Sigma = \left( \begin{array}{cc} S & s \\ s^T & \sigma \end{array} \right),
\]
where $Y,S \in \reals^{(n-1) \times (n-1)}$, $y,s\in\reals^{n-1}$, and $\myxi,\sigma \in \reals$. We are considering the problem above, where $Y$ is fixed, and $(y,\myxi) \in \reals^n$ is the variable. We use the notation $t := \Tr Y$.

The conic constraint $X \succ 0$ translates as $y^TY^\dagger y \le \myxi$, $y \in {\cal R}(Y)$, where ${\cal R}(Y)$ is the range of the matrix $Y$. We obtain the sub-problem
\begin{equation}\label{eq:sub-pb}
\psi := \dsp\max_{x,y} \: 
\left(\begin{array}{c}
2(y^Ts - \lambda \|y\|_1 ) + (\sigma-\lambda)\myxi - \frac{1}{2}(t+\myxi)^2 \\
+ \beta \log(\myxi - y^T Y^{\dagger} y) 
\end{array}\right)
~:~ y \in {\cal R}(Y).
\end{equation}
%The above is a convex problem that can be solved efficiently, in $O(n^3)$.  
\paragraph{Simplifying the sub-problem.} We can simplify the above problem, in particular, avoid the step of forming the pseudo-inverse of $Y$, by taking the dual of problem~(\ref{eq:sub-pb}).

Using the conjugate relation, valid for every $\eta >0$:
\[
\log \eta + 1 = \min_{z>0} \: z\eta - \log z,
\]
and with $f(x) := (\sigma -\lambda)x - {\onehalf}(t+x)^2$, we obtain
\begin{eqnarray*}
\psi+\beta &=& \max_{y \in {\cal R}(Y)} \: 2(y^Ts - \lambda\|y\|_1) + f(x) + \beta \min_{z>0} \: \left( z(x-y^TY^{\dagger}y) - \log z\right) \\
&=&  \min_{z>0} \: \max_{y \in {\cal R}(Y)} \: 2(y^Ts - \lambda\|y\|_1 -\beta zy^TY^{\dagger}y) + \max_x \: (f(x) +\beta zx)  - \beta \log z \\
& =& \min_{z>0} h(z) + 2 g(z)
\end{eqnarray*}
where, for $z>0$, we define
\begin{eqnarray*}
h(z) &:=& -\beta \log z + \max_x \: (f(x) + \beta z x) \\
&=&  -\beta \log z + \max_x \: ((\sigma -\lambda + \beta z)x - {\onehalf}(t+x)^2) \\
&=& -{\onehalf}t^2 -\beta \log z + \max_x \: ((\sigma -\lambda -t + \beta z)x - {\onehalf}x^2) \\
&=& -{\onehalf}t^2 -\beta \log z + \onehalf (\sigma -\lambda -t + \beta z)^2 
\end{eqnarray*}
with the following relationship at optimum:
\begin{equation}\label{eq:x-opt}
x =  \sigma -\lambda -t + \beta z.
\end{equation}
In addition,
\begin{eqnarray*}
g(z) &:=& \max_{y \in {\cal R}(Y)}  \: y^Ts - \lambda \|y\|_1 - \frac{\beta z}{2}( y^T Y^{\dagger} y)  \\
&=& \max_{y \in {\cal R}(Y)}  \: y^Ts  + \min_{v \::\: \|v\|_\infty \le \lambda}  y^Tv- \frac{\beta z}{2}( y^T Y^{\dagger} y)  \\
&=& \min_{v \::\: \|v\|_\infty \le \lambda} \: \max_{y \in {\cal R}(Y)} \: ( y^T(s+v) - \frac{\beta z}{2} (y^T Y^\dagger y) ) \\
&=& \min_{u \::\: \|u -s\|_\infty \le \lambda} \: \max_{y \in {\cal R}(Y)} \: ( y^Tu - \frac{\beta z}{2} (y^T Y^\dagger y) ) \\
&=& \min_{u \::\: \| u -s \|_\infty \le \lambda} \: \frac{1}{2\beta z} u^TY u.
\end{eqnarray*}
with the following relationship at optimum:
\begin{equation}\label{eq:y-opt}
y = \frac{1}{\beta z} Y u.
\end{equation}

Putting all this together, we obtain the dual of problem~(\ref{eq:sub-pb}): with $\psi' := \psi + \beta +{\onehalf}t^2$, and $ c:=\sigma -\lambda -t$, 
we have
\[
\psi' = \min_{u,z} \: \frac{1}{\beta z} u^TY u -\beta \log z + \onehalf (c+ \beta z )^2  ~:~ z>0, \;\; \|u-s\|_\infty \le \lambda.
\]
Since $\beta$ is small, we can avoid large numbers in the above, with the change of variable $\tau = \beta z$:
\begin{equation}\label{eq:dual-sub-pb}
\psi' -\beta\log \beta = \min_{u,\tau} \: \frac{1}{\tau} u^TY u -\beta \log \tau + \onehalf (c+ \tau)^2  ~:~ \tau>0, \;\; \|u-s\|_\infty \le \lambda.
\end{equation}
%The above problem is convex, and involves rotated second-order cone constraints. It can be solved by off-the-shelf softwares such as mosek using interior-point methods with a complexity of $O(n^3)$. One problem is that mosek requires to form the square-root of $Y$.
\paragraph{Solving the sub-problem.}  
Problem~(\ref{eq:dual-sub-pb}) can be further decomposed into two stages. 

First, we solve the box-constrained QP
\begin{equation}\label{eq:dual-sub-pb-qp}
R^2 := \dsp\min_{u} \: u^TY u ~:~  \|u-s\|_\infty \le \lambda,
\end{equation}
%is amenable to a simple interior-point method that exploits sparsity of $Y$. 
using a simple coordinate descent algorithm to exploit sparsity of $Y$. Without loss of generality, we consider the problem of updating the first coordinate of $u$. Partition $u$, $Y$ and $s$ as
\[
u = \left( \begin{array}{c} \eta \\ \hat u  \end{array} \right), \;\; Y = \left( \begin{array}{cc}  y_1 & \hat y^T \\ \hat y & \hat Y \end{array} \right), \;\;
s = \left( \begin{array}{c} s_1 \\ \hat s \end{array} \right),
\]
Where, $\hat Y \in \reals^{(n-2) \times (n-2)}$, $\hat u, \hat y, \hat s \in \reals^{n-2}$, $y_1, s_1 \in \reals$ are all fixed, while  $\eta \in \reals$ is the variable.
We obtain the subproblem 
\begin{equation}\label{eq:sub-sub-pb}
\min_\eta \: y_1 \eta^2 + (2 \hat y^T \hat u) \eta ~:~ \|\eta - s_1\| \le \lambda
\end{equation}
for which we can solve for $\eta$ analytically using the formula given below.
\begin{equation} \label{eq:box-qp-cd}
\eta = \left\{ \begin{array}{ll} -\frac{\hat y^T \hat u}{y_1} &  \textrm{if } \| s_1 + \frac{\hat y^T \hat u}{y_1} \| \le \lambda, \;y_1 > 0, \\
s_1 - \lambda & \textrm{if } -\frac{\hat y^T \hat u}{y_1} < s_1 - \lambda,\; y_1 >0 \textrm{ or if }  \hat y^T \hat u >0, \; y_1= 0, \\
s_1 + \lambda & \textrm{if } -\frac{\hat y^T \hat u}{y_1} > s_1 + \lambda,\; y_1 >0 \textrm{ or if } \hat y^T \hat u <=0,\; y_1= 0.
\end{array}
\right.
\end{equation}

Next, we set $\tau$ by solving the one-dimensional problem:  
\[
\dsp\min_{\tau>0} \: \dsp\frac{R^2}{\tau}  -\beta \log \tau + \onehalf (c+ \tau)^2 .
\]
The above can be reduced to a bisection problem over $\tau$, or by solving a polynomial equation of degree $3$. 

\paragraph{Obtaining the primal variables.} Once the above problem is solved, we can obtain the primal variables $y,x$, as follows. Using formula~(\ref{eq:y-opt}), with $\beta z = \tau$, we set $y = \frac{1}{\tau} Y u$.
For the diagonal element $x$, we use formula~(\ref{eq:x-opt}): $x  = c + \tau =\sigma -\lambda -t + \tau$.

\paragraph{Algorithm summary.} 
We summarize the above derivations in Algorithm~\ref{alg:bcd}. Notation: for any symmetric matrix $A \in \reals^{n \times n}$, let $A_{\backslash i \backslash j}$ denote the matrix produced by removing row $i$ and column $j$. Let $A_j$ denote column $j$ (or row $j$) with the diagonal element $A_{jj}$ removed.
\begin{algorithm}[t]
{\footnotesize
\caption{Block Coordinate Ascent Algorithm}
\label{alg:bcd}
\begin{algorithmic} [1]
\REQUIRE The covariance matrix $\Sigma$, and a parameter $\rho>0$. 
\STATE Set $X^{(0)} = I$
\REPEAT
\FOR{$j=1$ to $n$}  
\STATE Let $X^{(j-1)}$ denote the current iterate. Solve the box-constrained quadratic program
 \[
R^2 := \min_{u} \: u^T X_{\backslash j \backslash j}^{(j-1)}u ~:~  \|u-\Sigma_j\|_\infty \le \lambda
\] 
using the coordinate descent algorithm
\STATE Solve the one-dimensional problem
\[
\min_{\tau>0} \: \frac{R^2}{\tau}  -\beta \log \tau + \onehalf (\Sigma_{jj} -\lambda - \Tr X_{\backslash j \backslash j}^{(j-1)}+ \tau)^2 
\]
using a bisection method, or by solving a polynomial equation of degree $3$. 

\STATE First set $X_{\backslash j \backslash j}^{(j)} =X_{\backslash j \backslash j}^{(j-1)}$, and then set both $X^{(j)}$'s column $j$ and row $j$ using
\[
X_j^{(j)} =  \frac{1}{\tau} X_{\backslash j \backslash j}^{(j-1)} u
\]
\[
X_{jj}^{(j)} = \Sigma_{jj} -\lambda - \Tr X_{\backslash j \backslash j}^{(j-1)}+ \tau
\]
\ENDFOR
\STATE Set $X^{(0)} = X^{(n)}$
\UNTIL{convergence}
\end{algorithmic} 
}
\end{algorithm} 

\paragraph{Convergence and complexity.}
Our algorithm solves DSPCA by first casting it to problem~(\ref{eq:primal-pb-aug-pen}), which is in the general form~(\ref{eq:general-form}).
%which is in the form 
%\[
%\min_X \: f(X) - \beta \log \det X ~:~ L \le X \le U, \;\; X \succ 0,
%\]
%where $X = X^T$ is a $n \times n$ matrix variable, $L,U$ impose component-wise bounds on $X$, $f$ is convex, and $\beta >0$. 
Therefore, the convergence result from~\cite{WGMS:09} readily applies and hence every limit point that our block coordinate ascent algorithm converges to is the global optimizer. The simple coordinate descent algorithm solving problem~(\ref{eq:dual-sub-pb-qp}) only involves a vector product and can take sparsity in $Y$ easily. To update each column/row takes $O(n^2)$ and there are $n$ such columns/rows in total. Therefore, our algorithm has a computational complexity of  $O(Kn^3)$, where $K$ is the number of sweeps through columns.  In practice, $K$ is fixed at a number independent of problem size (typically $K = 5$). Hence our algorithm has better dependence on the problem size compared to $O(n^4\sqrt{\log n})$ required of the first order algorithm developed in~\cite{AEJL:07}. 

\begin{figure}[htb]
\begin{center}
\begin{tabular}{cc}
\includegraphics[width = 0.48\textwidth, height = 0.38\textwidth]{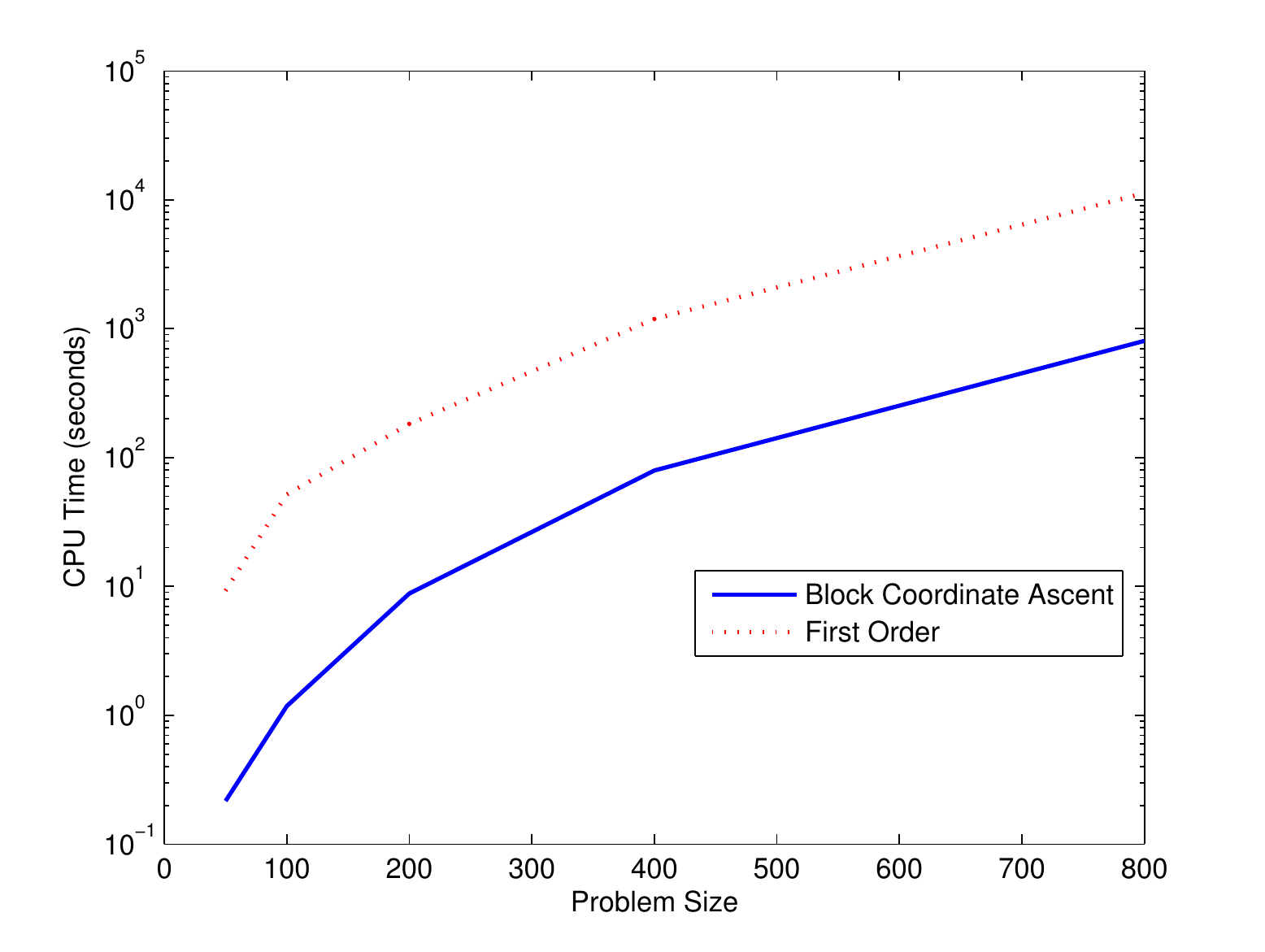}&
\includegraphics[width = 0.48\textwidth, height = 0.38\textwidth]{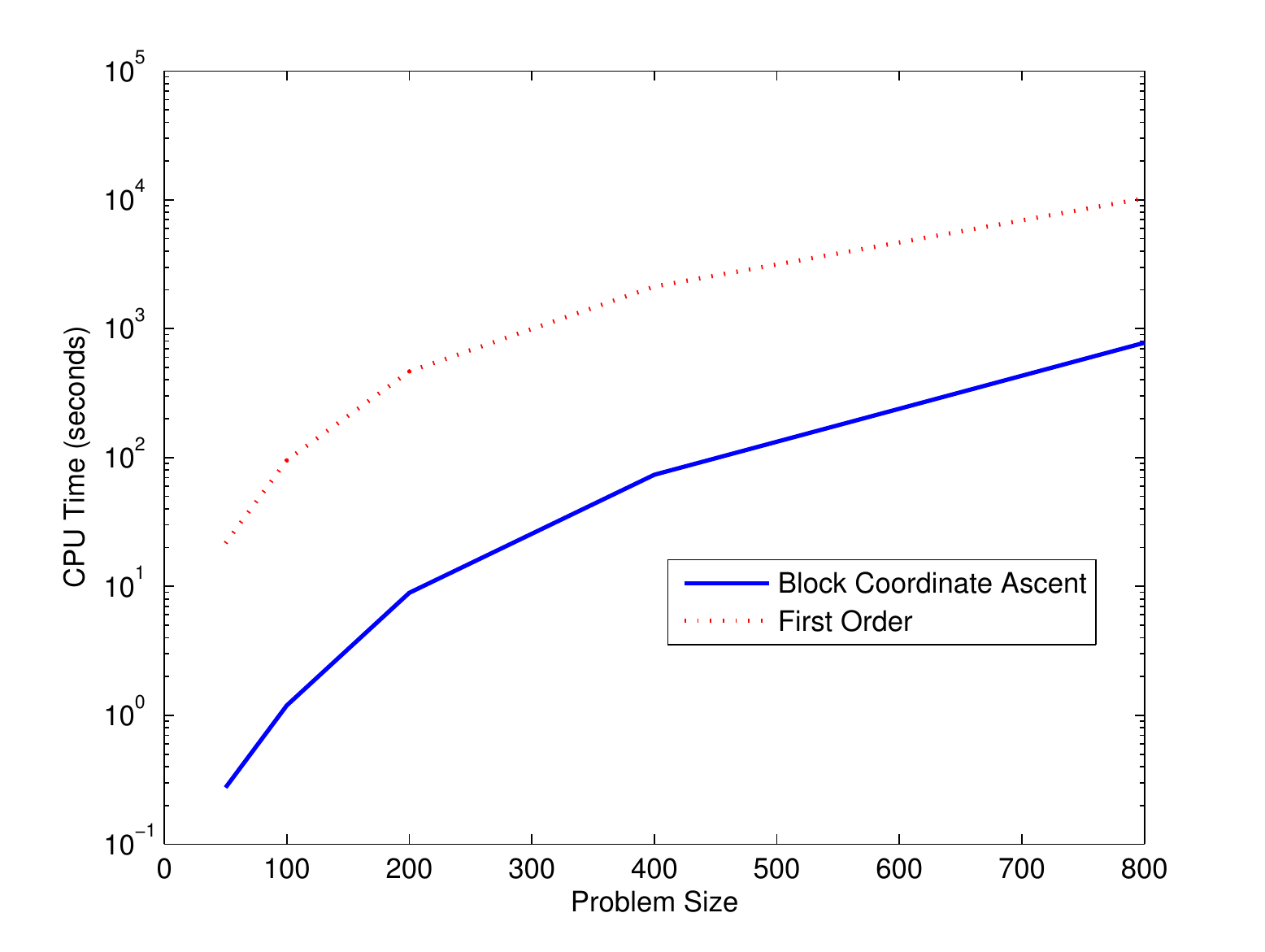}
\end{tabular}
\end{center}
\caption{Speed comparisons between Block Coordinate Ascent and First-Order\label{speed_cmp}}
\end{figure}

Fig~\ref{speed_cmp} shows that our algorithm converges much faster than the first order algorithm. On the left, both algorithms are run on a covariance matrix $\Sigma=F^TF$ with $F$ Gaussian. On the right, the covariance matrix comes from a "spiked model" similar to that in \cite{Amin08}, with $\Sigma = uu^T + VV^T/m$, where $u\in\reals^n$ is the true sparse leading eigenvector, with $\Card(u)= 0.1 n $, $V\in\reals^{n \times m}$ is a noise matrix with $V_{ij}\sim{\cal N}(0,1)$ and $m$ is the number of observations.
% where $k_1$ is the number of column sweeps for the outer block coordinate ascent algorithm and $k_2$ is the number of iterations for the coordinate descent algorithm to solve the inter box-constrained QP problem. 
%\paragraph{Memory issues.} 
%For the block coordinate ascent algorithm, we only need to keep one column of $\Sigma$ in memory, while the first-order algorithm in~\cite{AEJL:07} has to retain the whole $\Sigma$ (which is typically dense) in memory. Though both algorithms need to hold the intermediate solution in memory during iterations, that matrix is typically sparse when $\lambda$ is high. True or not? Need more work here.

\section{Numerical Examples}
\begin{figure}[h]
\begin{center}
\begin{tabular}{cc}
\includegraphics[width = 0.48\textwidth, height = 0.38\textwidth]{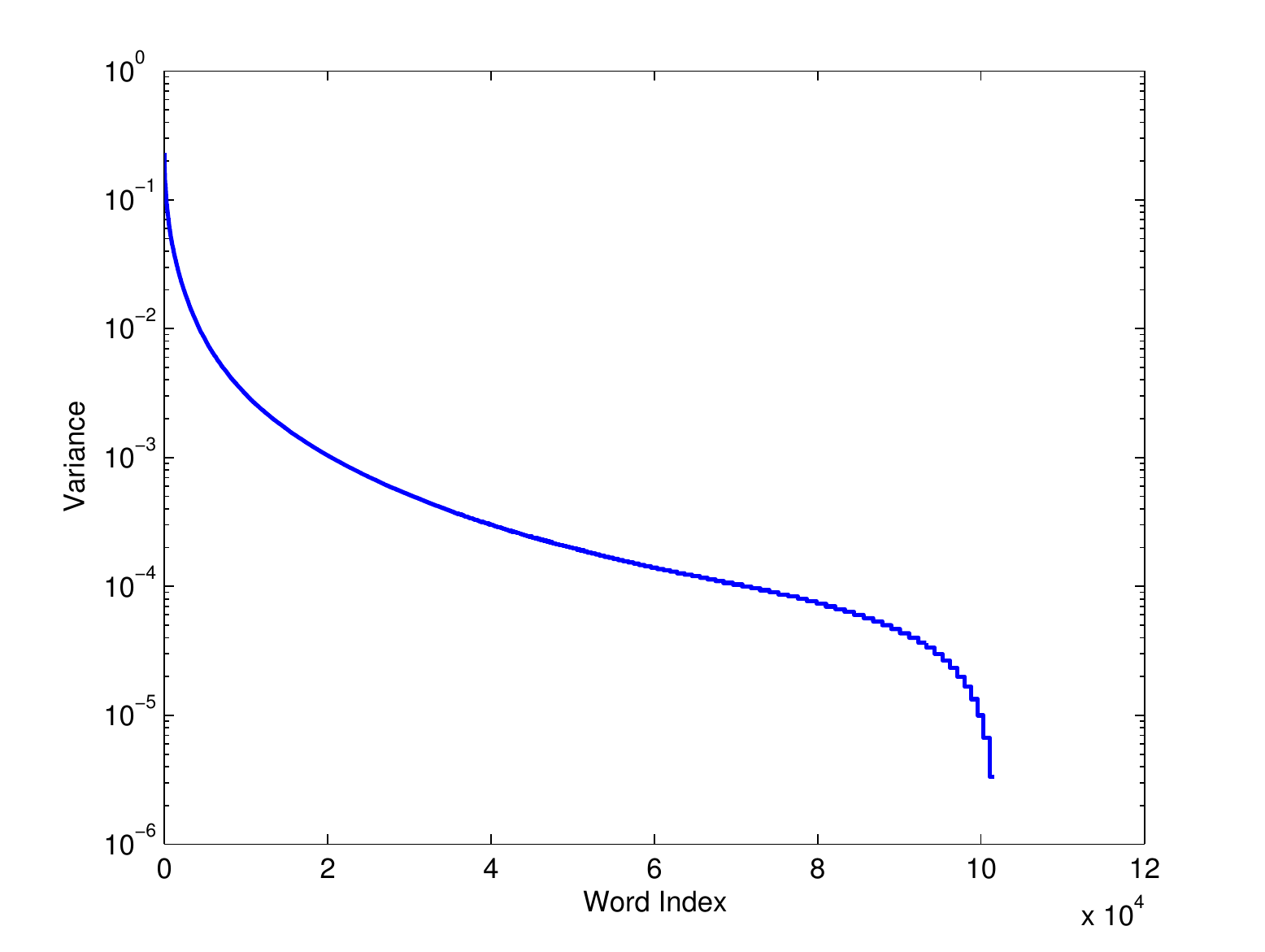}&
\includegraphics[width = 0.48\textwidth, height = 0.38\textwidth]{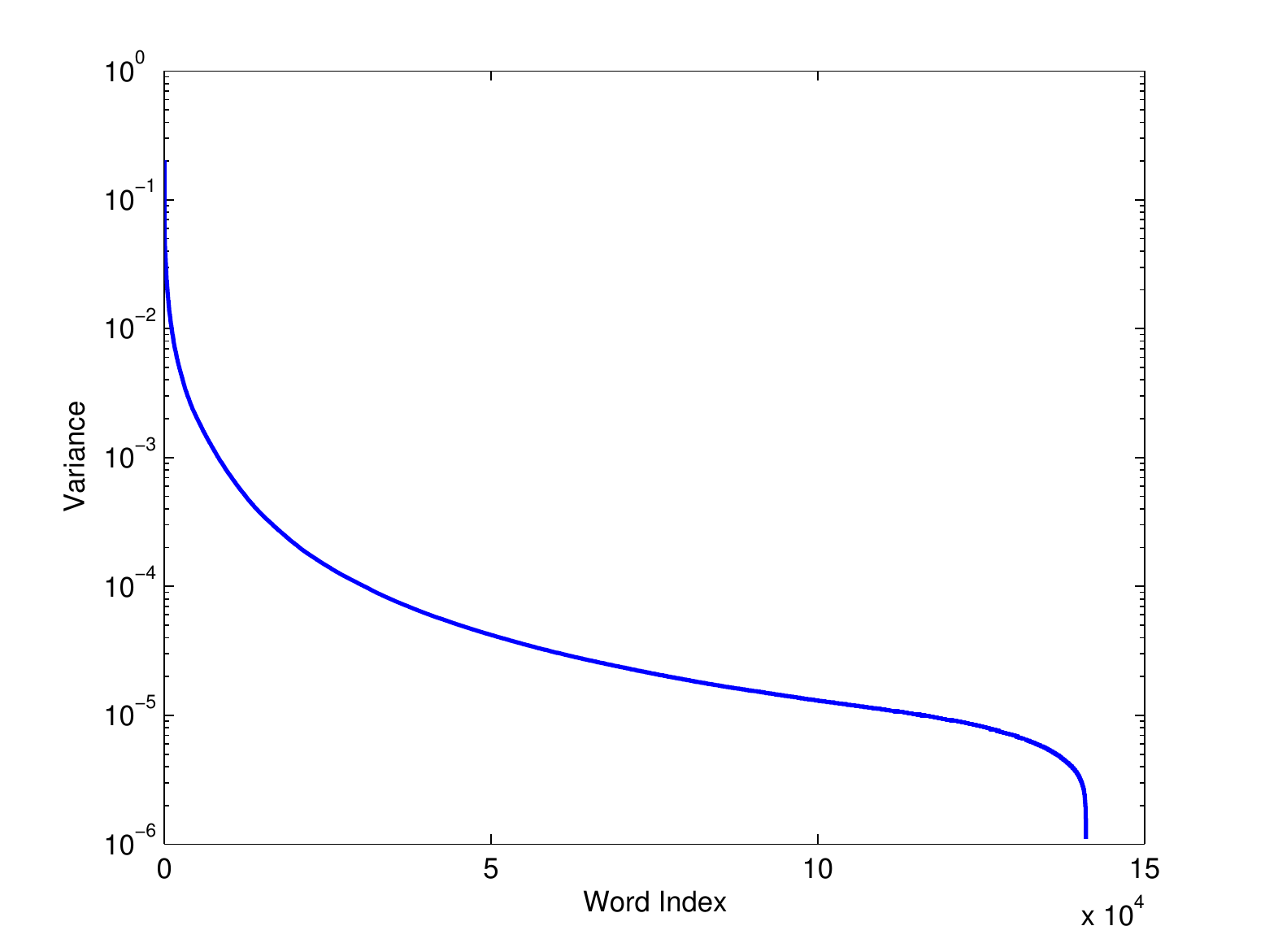}
\end{tabular}
\end{center}
\caption{ Sorted variances of 102,660 words in NYTimes (left) and 141,043 words in PubMed (right)\label{srtvar}}
\end{figure}

%In this section, we analyze two publicly available large data sets: NYTimes news articles and PubMed abstracts, available from UCI Machine Learning Repository \cite{FA:10}. Both text collections are in the form of bag-of-words. NYTtimes text collection contains $300,000$ articles and has a dictionary of $102,660$ unique words, resulting a file of size 1 GB. The even larger PubMed data has $8,200,000$ abstracts with $141,043$ unique words in them, giving a file of size 7.8 GB. These data are so large that we cannot even load them into memory all at once and hence cannot form a whole data matrix (not mention to form covariance matrix) in memory for a classical PCA analysis. However with the preprocessing technique presented in Section 2 and the block coordinate ascent algorithm developed in Section 3, we are able to perform sparse PCA analysis of these data, also thanks to the fact that variances of words decrease drastically when we rank them as shown in Fig~\ref{srtvar}. 
In this section, we analyze two publicly available large data sets, the NYTimes news articles data and the PubMed abstracts data, available from the UCI Machine Learning Repository \cite{FA:10}. Both text collections record word occurrences in the form of bag-of-words. The NYTtimes text collection contains $300,000$ articles and has a dictionary of $102,660$ unique words, resulting in a file of size 1 GB. The even larger PubMed data set has $8,200,000$ abstracts with $141,043$ unique words in them, giving a file of size 7.8 GB. These data matrices are so large that we cannot even load them into memory all at once, which makes even the use of classical PCA difficult. However with the pre-processing technique presented in Section 2 and the block coordinate ascent algorithm developed in Section 3, we are able to perform sparse PCA analysis of these data, also thanks to the fact that variances of words decrease drastically when we rank them as shown in Fig~\ref{srtvar}.  Note that the feature elimination result only requires the computation of each feature's variance, and that this task is easy to parallelize.

By doing sparse PCA analysis of these text data, we hope to find interpretable principal components that can be used to summarize and explore the large corpora. Therefore, we set the target cardinality for each principal component to be $5$. As we run our algorithm with a coarse range of $\lambda$ to search for a solution with the given cardinality, we might end up accepting a solution with cardinality close, but not necessarily equal to, $5$, and stop there to save computational time. 

The top $5$ sparse principal components are shown in Table~\ref{tab:nyt-spc} for NYTimes and in Table~\ref{tab:pubmed-spc} for PubMed. Clearly the first principal component for NYTimes is about business, the second one about sports, the third about U.S., the fourth about politics and the fifth about education. Bear in mind that the NYTimes data from UCI Machine Learning Repository ``have no class labels, and for copyright reasons no filenames or other document-level metadata'' \cite{FA:10}. The sparse principal components still unambiguously identify and perfectly correspond to the topics used by \emph{The New York Times} itself to classify articles on its own website. %{\tiny 
\begin{table}[htb]
\centering 
\caption{\label{tab:nyt-spc} Words associated with the top 5 sparse principal components in NYTimes}
%\texttt{
{\small
\begin{tabular}{lllll}
  {\rm 1st PC (6 words)} & {\rm 2nd PC (5 words)} & {\rm 3rd PC (5 words)} & {\rm 4th PC (4 words)}& {\rm 5th PC (4 words)}\\
\hline
\texttt{million}&\texttt{point}&\texttt{official}&\texttt{president}&\texttt{school}\\
\texttt{percent}&\texttt{play}&\texttt{government}&\texttt{campaign}&\texttt{program}\\
\texttt{business}&\texttt{team}&\texttt{united\_states}&\texttt{bush}&\texttt{children}\\
\texttt{company}&\texttt{season}&\texttt{u\_s}&\texttt{administration}&\texttt{student}\\
\texttt{market}&\texttt{game}&\texttt{attack}&&\\
\texttt{companies}&&&&\\
\hline
\end{tabular}
%}
}
\end{table}
%}

After the pre-processing steps, it takes our algorithm around 20 seconds to search for a range of $\lambda$ and find one sparse principal component with the target cardinality (for the NYTimes data in our current implementation on a MacBook laptop with 2.4 GHz Intel Core 2 Duo processor and 2 GB memory).
\begin{table}[htb]
\centering 
\caption{\label{tab:pubmed-spc} Words associated with the top 5 sparse principal components in PubMed}
%\texttt{
{\small
\begin{tabular}{lllll}
  {\rm 1st PC (5 words)} & {\rm 2nd PC (5 words)} & {\rm 3rd PC (5 words)} & {\rm 4th PC (4 words)}& {\rm 5th PC (4 words)}\\
\hline
\texttt{patient}&\texttt{effect}&\texttt{human}&\texttt{tumor}&\texttt{year}\\
\texttt{cell}&\texttt{level}&\texttt{expression}&\texttt{mice}&\texttt{infection}\\
\texttt{treatment}&\texttt{activity}&\texttt{receptor}&\texttt{cancer}&\texttt{age}\\
\texttt{protein}&\texttt{concentration}&\texttt{binding}&\texttt{maligant}&\texttt{children}\\
\texttt{disease}&\texttt{rat}&&\texttt{carcinoma}&\texttt{child}\\
\hline
\end{tabular}
%}
}
\end{table}

A surprising finding is that the safe feature elimination test, combined with the fact that word variances decrease rapidly, enables our block coordinate ascent algorithm to work on covariance matrices of order at most $n=500$, instead of the full order ($n=102660$) covariance matrix for NYTimes, so as to find a solution with cardinality of around $5$. In the case of PubMed, our algorithm only needs to work on covariance matrices of order at most $n=1000$, instead of the full order ($n=141,043$) covariance matrix. Thus, at values of the penalty parameter $\lambda$ that target cardinality of $5$ commands, we observe a dramatic reduction in problem sizes, about $150\sim 200$ times smaller than the original sizes respectively. This motivates our conclusion that sparse PCA is in a sense, easier than PCA itself. 
%{\color{red} To do if time allows:
%\begin{itemize}
%\item comparison with Shen and Huang's method and GPower methods: for example, show words selected by these methods
%\end{itemize}
%}
%The PubMed data set contains similar number of features (141,043 words) to NYTimes, but has almost $30$ times as many samples (8,200,000 abstracts). This huge sample size presents another challenge of computing covariance matrix. Even if we apply the safe feature test and reduce the feature size down to 500, we still encounter memory problem when trying to calculating the covariance matrix. Luckily, it turns out that in sparse PCA setting there is a scaling of the feature size n, the sample size m and the cardinality k to guarantee statistical consistency \cite{Amin08}.( review that result here.)  As a result, we can apply downsampling to reduce sample size.
\section{Conclusion}
The safe feature elimination result, coupled with a fast block coordinate ascent algorithm, allows to solve sparse PCA problems for very large scale, real-life data sets. The overall method works especially well when the target cardinality of the result is small, which is often the case in applications where interpretability by a human is key. The algorithm we proposed has better computational complexity, and in practice converges much faster than, the first-order algorithm developed in~\cite{AEJL:07}. Our experiments on text data also show that the sparse PCA can be a promising approach towards summarizing and organizing a large text corpus. 

%We also code our method into a software package available at (link here) for public use. 

\bibliographystyle{unsrt}
\bibliography{MainPerso}
\end{document}

%% file: defs.tex
\usepackage{algorithmic,algorithm}

\newcommand{\BEAS}{\begin{eqnarray*}}
\newcommand{\EEAS}{\end{eqnarray*}}
\newcommand{\BEA}{\begin{eqnarray}}
\newcommand{\EEA}{\end{eqnarray}}
\newcommand{\BEQ}{\begin{equation}}
\newcommand{\EEQ}{\end{equation}}
\newcommand{\BIT}{\begin{itemize}}
\newcommand{\EIT}{\end{itemize}}
\newcommand{\BNUM}{\begin{enumerate}}
\newcommand{\ENUM}{\end{enumerate}}

\newcommand{\BA}{\begin{array}}
\newcommand{\EA}{\end{array}}

\newcommand{\reals}{{\mbox{\bf R}}}

\newcommand{\Card}{\mathop{\bf Card}}
\newcommand{\Tr}{\mathop{\bf Tr}}

\newcommand{\dsp}{\displaystyle}

\newcounter{exno}

{\begin{quote}}{\end{quote}}

\makeatletter
\long\def\@makecaption#1#2{
   \vskip 9pt 
   \begin{small}
   \setbox\@tempboxa\hbox{{\bf #1:} #2}
   \ifdim \wd\@tempboxa > 5.5in
        \begin{center}
        \begin{minipage}[t]{5.5in}
        \addtolength{\baselineskip}{-0.95pt}
        {\bf #1:} #2 \par
        \addtolength{\baselineskip}{0.95pt}
        \end{minipage}
        \end{center}
   \else 
    \hbox to\hsize{\hfil\box\@tempboxa\hfil}  
   \fi
   \end{small}\par
}
\makeatother

\newcounter{oursection}

\newcounter{lecture}